\documentclass[letterpaper]{article}
\usepackage[preprint]{aaai2027}
\usepackage[hyphens]{url}
\usepackage{graphicx}
\usepackage{natbib}
\usepackage{caption}
\usepackage{booktabs}
\usepackage{colortbl}
\definecolor{cellY}{HTML}{C8E6C9}
\definecolor{cellP}{HTML}{FFF9C4}
\definecolor{cellN}{HTML}{FFCDD2}

\usepackage{amsmath}
\usepackage{xcolor}
\usepackage{listings}

\definecolor{codebg}{HTML}{F5F5F5}

\graphicspath{{figure/}}

\title{What Proves You Wrong: Benchmarking Language Models on Falsifiable Research Ideation}
\author{%
Ziyue Wang\equalcontrib\textsuperscript{1},
Aomufei Yuan\equalcontrib\textsuperscript{2},
Yiran Yao\equalcontrib\textsuperscript{3},
Linli Yao\textsuperscript{1},
Hongyao Zuo\textsuperscript{3},
Ziwen Gong\textsuperscript{4},
Yuanxin Liu\textsuperscript{1},
Shicheng Li\textsuperscript{1},
Yishuo Cai\textsuperscript{1},
Tong Yang\corresponding\textsuperscript{2},
Xu Sun\corresponding\textsuperscript{1},
Xiaohui Li\textsuperscript{5},
Haoli Bai\textsuperscript{5}%
}
\affiliations{%
\textsuperscript{1}State Key Laboratory of Multimedia Information Processing, School of Computer Science, Peking University \\
\textsuperscript{2}Peking University \quad
\textsuperscript{3}Tianjin University \quad
\textsuperscript{4}Hainan University \quad
\textsuperscript{5}Huawei Technologies \\
\texttt{\{zywang25@stu.pku.edu.cn, rrustleer@gmail.com\}}%
}

\begin{document}

\maketitle

\begin{abstract}
Large language models are increasingly used to propose research ideas, yet the prevailing ways of judging such ideas supply no shared decision rule: free-form judging sways with style and position, and scoring against a later paper rewards recovery of one realized trajectory. We introduce a benchmark that carries a proposal from \textbf{Lit}erature \textbf{to} \textbf{Test}: the \textbf{Lit2Test} benchmark centers on a six-field contract organized around a falsifying outcome, so that every proposal precommits the observation that would prove it wrong, making its quality decidable in the first place rather than merely arguable. Built prospectively from 200 real-paper neighborhoods, Lit2Test elicits proposals from four frontier models and compares them through 1{,}200 pairwise comparisons judged blind in both presentation orders. The protocol audits its own reliability through diagnostic controls and bounded human calibration, with three annotators corroborating the conclusions within explicitly stated reliability bounds. Lit2Test recovers a strict ranking of the four models in all 10{,}000 bootstrap replicates, and the separation comes from the quality of the proposed tests and metrics rather than from surface fluency. We release the benchmark, construction pipeline, and audit artifacts for public use.\footnote{Data and code: \url{https://github.com/rrustlee/lit2test-benchmark}}

\end{abstract}

\section{Introduction}

\begin{figure*}[t]
\centering
\includegraphics[width=0.815\textwidth]{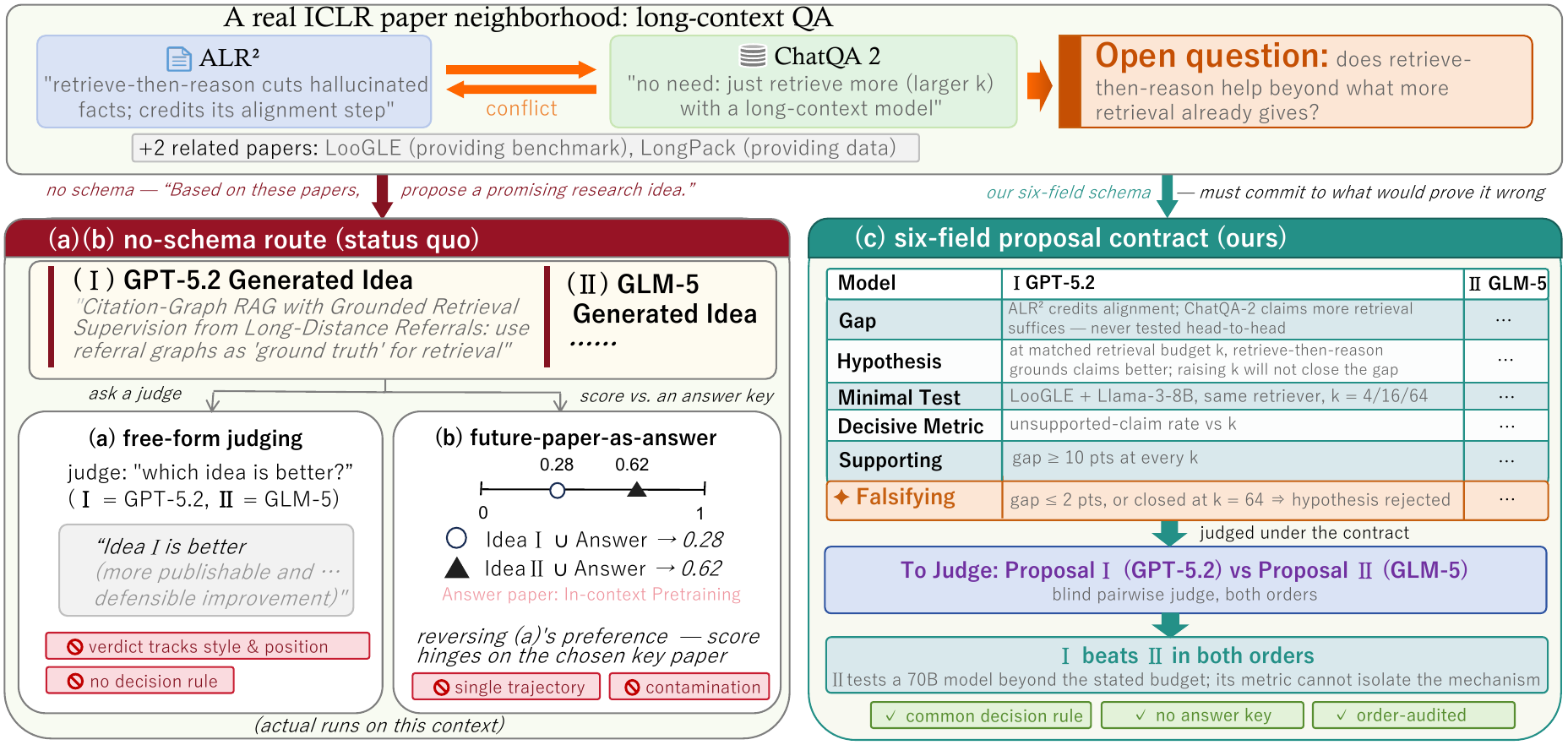}
\caption{Three evaluation paradigms on the same input, a real ICLR neighborhood on long-context QA; every panel shows actual runs on this context, with idea~I generated by GPT-5.2 and idea~II by GLM-5. The two existing paradigms disagree: (a)~free-form judging prefers idea~I in both presentation orders, while (b)~future-paper-as-answer scoring~\cite{shi2024incontext} prefers idea~II and reverses with the choice of answer-key paper. (c)~Our six-field contract route judges the same two models' proposals written under the contract and yields a grounded verdict; ``order-audited'' means each pair is judged in both presentation orders.}
\label{fig:paradigms}
\end{figure*}

A research proposal becomes actionable when it commits, in advance, to an observation that would force its own rejection. Without that commitment an idea can sound compelling while remaining untestable; with it, quality stops being a matter of taste and becomes something an experiment can decide. Figure~\ref{fig:paradigms} grounds this claim in a real four-paper ICLR neighborhood on long-context question answering~\cite{li2024canlongcontext,wang2024alr2,xu2025chatqa2,zhuang2025longdistance}, where two of the papers disagree: ALR\textsuperscript{2} reports that retrieve-then-reason cuts hallucinated facts and credits its own alignment step, while ChatQA~2 counters that retrieving more with a long-context model suffices. The neighborhood leaves an open question: does retrieve-then-reason help beyond what more retrieval already gives? A fluent, schema-free idea can speak to this tension without saying what measurement would settle it. A minimal falsifiable test instead specifies the controlled comparison, the decisive metric, and the outcome that would reject it; Figure~\ref{fig:paradigms}(c) renders this commitment as the orange row. Whether language models can reliably turn tension into test is the capability this paper measures.

Measuring this capability requires an instrument that survives its own audit, and the two existing paradigms fail in sequence. Free-form judging directly asks an LLM judge which of two ideas is better; on this neighborhood the verdict tracks style and presentation position (Figure~\ref{fig:paradigms}(a)), echoing known judge biases~\cite{zheng2023mtbench,wang2024faireval}, and ``idea quality'' is at root an object that gives the judge no decision rule. A second paradigm measures how well an idea anticipates a published follow-up paper~\cite{qiu2025aiideabench,guo2025ideabench}; it restores a reference signal yet fails differently: the score rewards alignment with one realized trajectory, penalizing valid alternatives that pursued a different direction, and invites knowledge-cutoff contamination whenever models can guess the target paper. The resulting instability is visible in Figure~\ref{fig:paradigms}(b), where the verdict conflicts with the free-form judge in (a) once the answer-key paper changes. These failures motivate a benchmark that carries a proposal from \textbf{Lit}erature \textbf{to} \textbf{Test}; Table~\ref{tab:related} (\S5) compares Lit2Test with its closest neighbors.

We ask three research questions about the models; each also asks whether the instrument itself can be trusted. \textbf{RQ1 (ordering and robustness):} what ordering of frontier models does Lit2Test produce, and is it robust to presentation order and resampling? \textbf{RQ2 (what drives the ordering):} which fields of the contract separate models, and does the judge respond to substance rather than style? \textbf{RQ3 (human corroboration):} do sampled humans corroborate the aggregate conclusions, and where does agreement break down?

Answering these questions imposes four design goals. \textbf{G1, literature grounding:} every instance is a real, provenance-tracked paper neighborhood rather than a bare topic. \textbf{G2, an executable common unit:} a six-field contract binds a literature gap, a hypothesis, a minimal test, a decisive metric, and bidirectional supporting and falsifying outcomes; its rubric rewards small-and-executable designs over grand-and-vague ones and leaves realized execution outcomes out of scope. \textbf{G3, prospective comparability:} all models receive the same fixed context, and no future paper is a privileged continuation. \textbf{G4, auditable measurement:} blind judgments in both orders, isolation of order-sensitive cases, ordinal statistics, controls, and human calibration jointly decide when a comparison is interpretable.

The Lit2Test benchmark realizes these goals at scale: 200 neighborhoods built from 800 unique papers in five batches, with four participant models (GPT-5.2, Claude Sonnet 4.6, GLM-5, and DeepSeek-V3.2) each contributing one six-field proposal per neighborhood, giving 1{,}200 canonical pairs and 2{,}400 blind ordered judgments by a non-participant judge.

Requiring every proposal to precommit its own refutation may look like a handicap on creativity; the requirement is not ours alone. A recent ideation system, ResearchStudio-Idea (IdeaSpark), imposes mechanism-linked falsification predictions as a generation-time safeguard, though its final quality judging does not score that prediction; Lit2Test instead makes the complete minimal-falsifiable-test contract the judged evaluation unit~\cite{zhao2026researchstudioidea}. Our own controls (\S4) confirm that the contract does not penalize quality: structured rendering is not by itself a gain, and contract-constrained proposals clearly beat naive baselines.

Our contributions:
\begin{itemize}
\setlength{\itemsep}{1pt}
\setlength{\parskip}{0pt}
\item A six-field \emph{joint evaluation unit} centered on minimal-falsifiable-test formulation, verified against a systematic survey of existing benchmarks (\S5; full survey in Appendix~F).
\item A \emph{prospective construction pipeline}: 200 real-paper neighborhoods and 800 model proposals built without future-paper answer keys, with provenance audits closing the contamination gap (\S2).
\item A \emph{reliability-audited protocol}: order-aware pairwise judgment with folded aggregation, controls, diagnostics, and bounded human calibration (\S3).
\item \emph{Empirical findings}: three bounded findings on ordering robustness, capability drivers, and human corroboration (\S4).
\end{itemize}

\section{Benchmark Construction}

We formalize the task and its unit (\S2.1--\S2.2), then construct the 200 instances (\S2.3--\S2.5), realizing design goals G1--G3; the protocol realizing G4 is defined in \S3.

\subsection{Task definition and scope}

An instance of the benchmark is a literature context $c$: a real four-paper neighborhood whose materials are fixed at construction time, chosen so that the papers share a topic while leaving a cross-paper tension that a small experiment could adjudicate. Every participant model receives the same $c$ and nothing else, so that differences in output reflect proposal quality rather than retrieval differences. Given $c$, a model must return a single proposal $P = (\textit{literature\_gap}$, \textit{hypothesis}, \textit{minimal\_test}, \textit{decisive\_metric}, \textit{supporting\_result}, $\textit{falsifying\_result})$.
The benchmark is the set of 200 contexts, and measurement compares proposal pairs $(P_i, P_j \mid c)$ through blind pairwise judgment (\S3). In scope are literature synthesis, gap identification, hypothesis formulation, and the design of an \emph{executable} minimal falsifiable test; execution feasibility is a first-class judged property. Out of scope are scientific creativity at large, forecasting of future publications, realized execution outcomes, and any claim that the workflow improves a model's generation: the task measures whether a proposal is testable under the current literature, not whether it will succeed.

\subsection{Six-field contract and design goals}

The six fields are the minimal contract under which a free-form idea becomes judgeable: remove any one, and a pairwise verdict loses its common decision structure. \textit{literature\_gap} ties the proposal to the supplied neighborhood so that grounded synthesis, not generic brainstorming, earns credit. \textit{hypothesis} sharpens a direction into a claim with conditions, mechanism, and an expected difference, so the judge adjudicates a statement, not an aspiration. \textit{minimal\_test} normalizes proposal granularity to avoid rewarding the largest agenda: proposals compete on how little suffices to discriminate the hypothesis, where the rubric's anti-grandiosity clause applies (G2). \textit{decisive\_metric} names the measurement able to adjudicate the mechanism, to avoid convenient aggregates under which every outcome reads as progress. \textit{supporting\_result} and \textit{falsifying\_result} together form a bidirectional decision rule: by precommitting both the confirming and the rejecting observation, they convert falsifiability from an abstract virtue into a checkable property of the text. A falsifier need not be numerical; a qualitative direction is falsifiable when the metric, comparison, and contradictory observation are operationally clear.

\subsection{Source selection and provenance}

The benchmark comprises 200 literature contexts drawn from 800 unique source publications, organized as five batches of 40 neighborhoods and assembled from recent OpenReview/ICLR-adjacent literature. Deduplication guarantees that no paper repeats across neighborhoods; source matching verifies that each neighborhood coheres around a shared question; and a provenance audit records the origin of every paper (G1). No target or future paper is designated as the correct continuation: the neighborhood supplies context, and there is no answer key (G3). Batch dates and seed naming are documented in the construction timeline of Appendix~B.

\subsection{Proposal generation}

Four participant models (GPT-5.2, Claude Sonnet 4.6, GLM-5, and DeepSeek-V3.2) each produce one native six-field proposal per neighborhood (generated directly in the schema, not converted afterwards) under a common prompt and shared generation settings, yielding 800 proposals. Schema validation checks all six fields on every response, with adjudication and bounded retries for malformed outputs, so that comparisons reflect proposal quality rather than prompt or interface differences.

\begin{figure*}[t]
\centering
\includegraphics[width=0.80\textwidth]{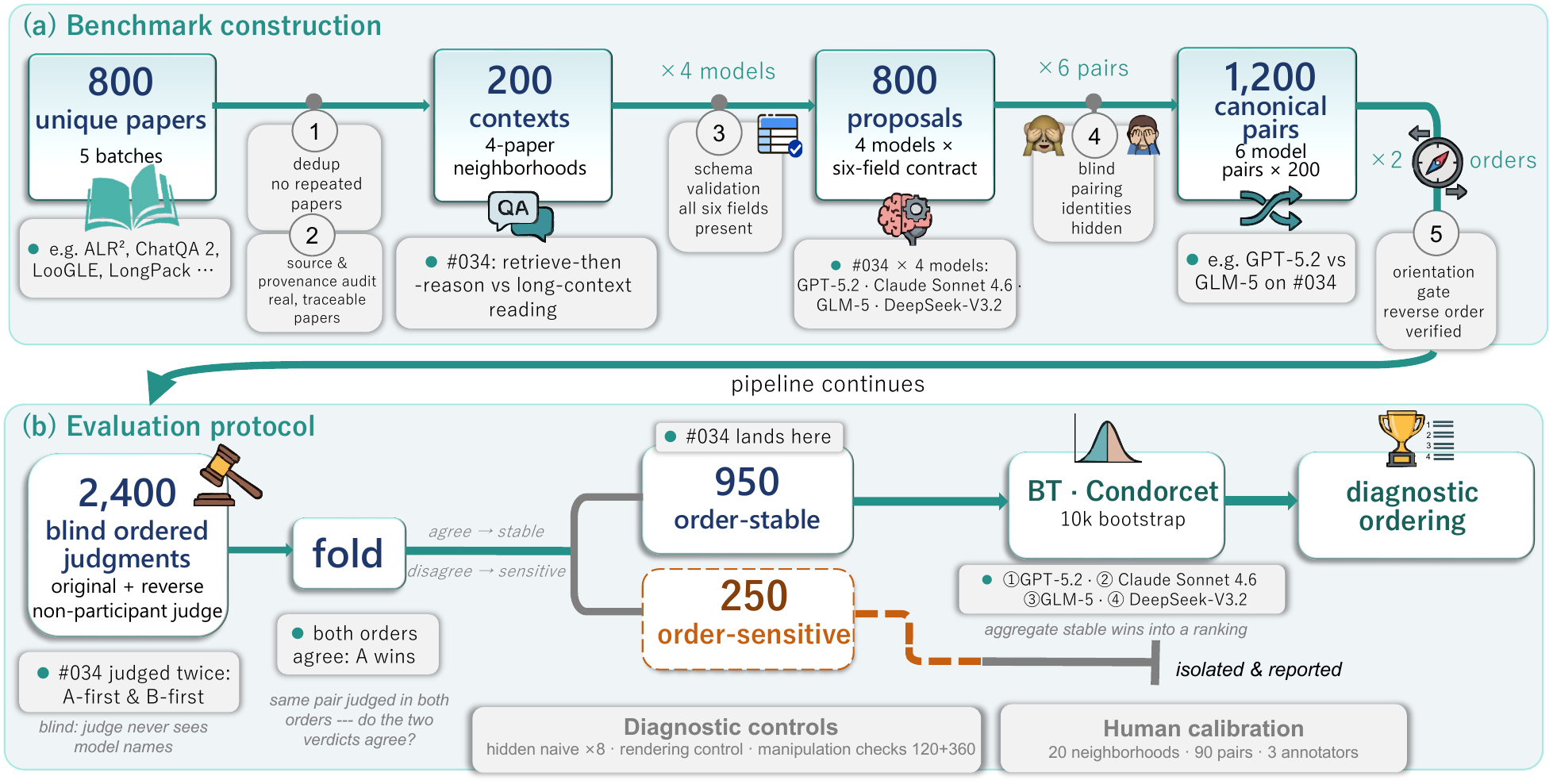}
\caption{The Lit2Test pipeline: (a)~benchmark construction and (b)~the shipped evaluation protocol, with specimen chips tracing neighborhood \#034 (the Figure~\ref{fig:paradigms} neighborhood) through every stage. Each of the 1{,}200 canonical pairs is judged in both orders and finally folded into 950 order-stable plus 250 order-sensitive cases; only stable cases feed the ranking aggregation.}
\label{fig:pipeline}
\end{figure*}

\subsection{Pair construction}

For each context, the four proposals form six model pairs, giving $200 \times 6 = 1{,}200$ canonical pairs (folded in \S3.2). Each canonical pair is judged in both presentation orders, original and reverse, under blind model identity, producing 2{,}400 ordered judgments; an orientation gate verifies for every reverse task that the A/B contents are genuinely swapped, so that the order-sensitivity analysis of \S3 rests on true reversals. Figure~\ref{fig:pipeline}(a) shows the construction band of this pipeline, following one specimen, neighborhood \#034 of Figure~\ref{fig:paradigms}, through every stage. A stratified subsample of neighborhoods later serves human calibration (\S4.3).

\section{Evaluation Protocol}

This protocol is part of the Lit2Test artifact: using the benchmark means running these judgments, this folding rule, and these controls. Figure~\ref{fig:pipeline}(b) traces the flow; the order-sensitive branch terminates in reporting and never rejoins the aggregation path. \S4 reports results.

\subsection{Blind pairwise judgment}

The canonical observable is a blind pairwise verdict~\cite{zheng2023mtbench,liu2023geval}: the judge, Gemini~3.1 Pro (Preview), fixed per benchmark version, receives two anonymized six-field proposals for the same neighborhood with an explicit rubric and returns A, B, or TIE. The judge never participates as a generator, removing self-preference within the participant set~\cite{zheng2023mtbench}. The holistic verdict is canonical because it matches the ordinal comparison the aggregation consumes; per-dimension scores would require cardinal, equal-weight treatment across heterogeneous rubric dimensions, so we collect them only as a diagnostic layer.

\subsection{Order folding}

Because pairwise judges exhibit position bias~\cite{wang2024faireval,zheng2023mtbench}, each canonical pair is judged in both presentation orders, and the two ordered verdicts are folded into one case outcome: a case is \emph{order-stable} when both orientations agree on a winner after accounting for side, and \emph{order-sensitive} otherwise: the winner reverses, or at least one orientation returns a tie. The order-sensitive set is an explicit deliverable reported alongside the ordering: a verdict that reverses under presentation marks a comparison the judge cannot resolve, and a symmetric average would erase that information. Throughout, the statistical unit is the canonical pair rather than the ordered row; treating forward and reverse judgments as independent observations would double the apparent sample.

\subsection{Aggregation and uncertainty}

Order-stable outcomes feed standard Bradley--Terry estimation, and Condorcet head-to-head relations summarize dominance without parametric assumptions. Uncertainty comes from a case-level bootstrap that resamples canonical cases (10{,}000 replicates), so reported stability reflects case-level rather than row-level variability. We fix in advance a policy for complete separation: when one model wins essentially every stable comparison against another, Bradley--Terry strengths diverge and their point values carry no calibrated meaning, so we report the ordinal structure (ranks, Condorcet relations, and bootstrap rank recovery) as the interpretable result and treat strength magnitudes as diagnostic only.

\subsection{Diagnostic controls}

A control family, defined here with results in \S4, probes whether the workflow measures the intended construct. A \emph{dimension-decomposed audit} re-judges a 90-pair subset with structured scores on five rubric dimensions (grounding, hypothesis specificity, minimality/feasibility, decisive metric, falsifiability) and checks their agreement with the canonical holistic verdict under both orders. \emph{Hidden controls} insert proposals from a naive keyword/template baseline blind among real pairs (eight in the automated audit, four more in the human study), doubling as a \emph{real-versus-naive anchor} the judge must stay above. A \emph{same-source rendering control} presents schema and prose renderings of identical content, separating substance from formatting. A \emph{clear single-field manipulation check} corrupts exactly one field (grounding, decisive metric, or falsifiability) with an obvious defect and requires the clean proposal to win in both orders. A \emph{subtle-corruption audit} replaces obvious defects with naturalistic ones, each paired against a style-matched sham edit, so preference for the clean proposal is measured net of surface rewriting. A bounded human calibration study complements these controls: three annotators label the stratified sample of \S2, with majority labels compared against the judge on decisive order-stable cases (\S4, Appendix~E).

\paragraph{Scope of claims.}
We report agreement between verdicts and reference signals, and reserve \emph{accuracy} for tasks with an objective label, such as hidden-control detection. Human labels are calibration evidence on a stratified subset; ground truth for open-ended proposals remains contested. The statistical unit is the canonical pair throughout. The results in \S4 characterize the reliability and construct validity of measurement on this benchmark; benchmark-wide human validation and the downstream experimental success of proposed tests lie beyond what this protocol establishes. The canonical judge and verdict rule are fixed per benchmark version, and any replacement requires a versioned full rerun.

\section{Experiments and Empirical Findings}

Results are organized by research question and, unless noted, cover all 1{,}200 canonical pairs and 2{,}400 ordered judgments under the protocol of \S3. Lit2Test deliberately reports few aggregate numbers: the object under test is a single capability and the reliability of its own measurement, not coverage across subtasks.




\begin{table*}[t]
\centering
\footnotesize
\setlength{\tabcolsep}{4pt}
\begin{tabular}{@{}l c cc cccc c@{}}
\toprule
& & \multicolumn{2}{c}{Stable W--L--T} & \multicolumn{4}{c}{Folded head-to-head (W--L--T)} & \\
\cmidrule(lr){3-4} \cmidrule(lr){5-8}
Model & BT [95\% CI] & Primary & J2 & vs.\ GPT & vs.\ CS & vs.\ GLM & vs.\ DS & Human WR \\
\midrule
GPT-5.2           & $\phantom{-}1.26$ [$\phantom{-}1.14,\phantom{-}1.40$] & 424--49--127  & 423--52--125  & ---          & 90--38--72  & 154--7--39  & 180--4--16  & .805 \\
Claude Sonnet 4.6 & $\phantom{-}0.74$ [$\phantom{-}0.62,\phantom{-}0.87$] & 347--119--134 & 391--74--135  & 38--90--72   & ---         & 143--18--39 & 166--11--23 & .775 \\
GLM-5             & $-0.73$ [$-0.86,-0.61$]                               & 118--343--139 & 97--405--98   & 7--154--39   & 18--143--39 & ---         & 93--46--61  & .233 \\
DeepSeek-V3.2     & $-1.28$ [$-1.42,-1.15$]                               & 61--439--100  & 69--449--82   & 4--180--16   & 11--166--23 & 46--93--61  & ---         & .214 \\
\bottomrule
\end{tabular}
\caption{Main results on Lit2Test.
\textbf{BT}: centered log-ability from the Bradley--Terry fit on order-stable cases (\S3.3); 95\% CIs from a 10{,}000-replicate case-level bootstrap.
\textbf{Primary / J2 Stable W--L--T}: per-model wins--losses--ties over the 600 folded cases involving that model; W and L count order-stable cases, T counts order-sensitive cases. The primary judge is Gemini~3.1 Pro (Preview) (950 stable / 250 sensitive); J2 is an independent second judge (Doubao Seed 2.0 Pro) re-judging all 2{,}400 comparisons (980 stable / 220 sensitive).
\textbf{Folded head-to-head}: per-pair W--L--T from the row model's perspective over 200 folded cases per pair (primary judge).
\textbf{Human WR}: majority win rate from the stratified human-calibration study (20 neighborhoods, 90 real pairs, 3 annotators; ties counted as 0.5).
Abbreviations: GPT = GPT-5.2, CS = Claude Sonnet 4.6, GLM = GLM-5, DS = DeepSeek-V3.2.}
\label{tab:main}
\end{table*}

\subsection{RQ1: What ordering, and how robust}

Folding the 2{,}400 ordered judgments into 1{,}200 canonical cases yields 950 order-stable and 250 order-sensitive cases. On the stable signal, Bradley--Terry estimation and Condorcet analysis both recover the ordering
\[
\footnotesize
\text{GPT-5.2} > \text{Claude Sonnet 4.6} > \text{GLM-5} > \text{DeepSeek-V3.2},
\]
and this full ordering is recovered in all 10{,}000 case-level bootstrap replicates (Table~\ref{tab:main}). A context-cluster bootstrap that resamples the 200 neighborhoods rather than the 1{,}200 pairs yields 11--27\% wider confidence intervals but recovers the identical ranking in all 10{,}000 replicates (Appendix~C).
The ordering is a strict Condorcet order: each higher-ranked model wins its stable head-to-head against every lower-ranked model, and it is consistent across all five construction batches. The 250 order-sensitive cases are retained and reported separately; keeping them out of the aggregation ensures the reported ordering is free of position artifacts. Two robustness checks bound this ordering further. Scoring all 250 order-sensitive cases as ties leaves the ordering unchanged with full bootstrap recovery, and even an adversarial assignment of every sensitive case preserves the two-tier structure (GPT-5.2 and Claude Sonnet 4.6 above GLM-5 and DeepSeek-V3.2), though within-tier adjacencies can flip. A second judge (Doubao Seed 2.0 Pro), from a model family disjoint from all four participants and the primary judge, independently reproduces the identical ordering, agreeing with the primary judge on 86.1\% of order-stable pairwise comparisons (Appendix~E). Six-pair folded detail, flip-tie and complete-separation expansions, per-batch tables, and the tie-sensitivity analysis appear in Appendix~C.

\paragraph{Finding 1.} Lit2Test produces a strict, fully bootstrap-recovered model ordering while explicitly isolating order-sensitive cases (250 of 1{,}200) rather than hiding them, so the ordering carries a bounded reliability envelope rather than an unqualified leaderboard claim.

\subsection{RQ2: What drives the ordering}

We next ask which fields of the contract separate models, and whether the judge responds to substance. All 188 dimension-audit judgments are valid. A structured overall verdict agrees with the canonical holistic verdict on 152/180 ordered judgments (84.4\%, case-clustered 95\% CI 78.9--89.4\%), and on the eight hidden real-versus-naive controls every structured aggregation selects the real answer (8/8). Among natural model outputs, minimality/feasibility provides the strongest dimension-level diagnostic signal, whereas falsifiability yields non-tied dimension verdicts in only 33/180 judgments and is the sole decisive dimension in 1/180 comparisons. This low natural variance does not indicate an inert dimension: both manipulation checks below show that the judge enforces falsifiability when it is degraded, so natural proposals are clearing the gate rather than escaping it. Two controls address the style objection directly: a same-source rendering control shows that schema-versus-prose rendering of identical content does not by itself move the verdict; and a falsifier on/off ablation (Appendix~D) shows that requiring the falsifier field is an auditability constraint rather than a generation-quality gain. Score-sum agreement variants appear in Appendix~D.

\begin{figure*}[t]
\centering
\includegraphics[width=\textwidth]{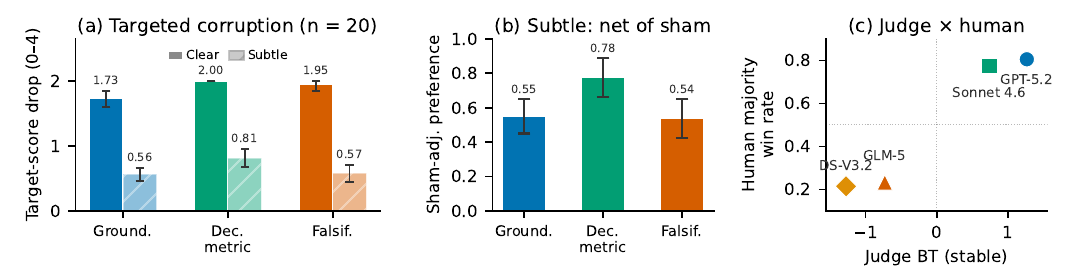}
\caption{Diagnostic controls and human calibration. (a)~Target-score drops under clear and subtle single-field corruptions (0--4 rubric, 95\% CIs; $n{=}20$ cases per dimension); clear corruptions produce large drops while subtle ones produce smaller but nonzero drops. (b)~Sham-adjusted preference for the clean proposal under subtle corruption, net of style-matched sham edits; all CIs exclude zero. (c)~Judge stable-case BT strength vs.\ human majority win rate; dashed lines mark the BT midpoint and the chance baseline. Both instruments recover the same ranking and the same two-tier separation. Human--judge agreement: 87.2\% on decisive stable cases (34/39), 11/12 naive-control detection, 88.3\% bootstrap rankings within one inversion.}
\label{fig:diagnostics}
\end{figure*}

The controlled single-field manipulation check verifies rubric responsiveness. When exactly one field (grounding, decisive metric, or falsifiability) is corrupted with a deliberately clear defect, clean proposals win all 40/40 ordered comparisons per dimension, with mean target-score drops of 1.73, 2.00, and 1.95 respectively and near-zero non-target drift (Figure~\ref{fig:diagnostics}a), confirming that each corruption penalizes only its intended field.

A contrast-controlled subtle-corruption audit extends this check to naturalistic defects. On the same 20 cases, each dimension receives naturalistic defect templates paired with style-matched sham edits; the sham-equivalence gate passes for all three dimensions (0 clean wins, 40 ties, 0 sham wins per dimension), so surface rewriting alone does not move the judge. Net of this sham baseline, the adjusted case-level preference for the clean proposal (defined in Appendix~D) is positive with 95\% CIs excluding zero for every dimension---0.550 [0.450, 0.650] for grounding, 0.775 [0.662, 0.887] for the decisive metric, and 0.537 [0.425, 0.650] for falsifiability, pooled 0.621 [0.567, 0.679] (Figure~\ref{fig:diagnostics}a--b)---while target-score drops attenuate to means of 0.56--0.81, versus 1.7--2.0 under clear corruption. This establishes contrast-controlled local sensitivity to naturalistic targeted defects under the frozen 20-case protocol; sham construction details and quality caveats appear in Appendix~D.

\paragraph{Finding 2.} Falsifiability defines an admissibility floor, while minimality/feasibility and decisive-metric design provide most observed separation among natural model outputs; the rendering control and both manipulation checks indicate that the judge tracks substance rather than style.

\subsection{RQ3: Human corroboration and its boundary}

A stratified calibration study tests whether sampled humans corroborate the aggregate conclusions: 20 neighborhoods covering 90 real model pairs and 4 hidden real-versus-naive controls, judged by 3 annotators blind to model identity. All three annotators are senior computer-science undergraduates working from a written bilingual protocol whose per-dimension rubric anchors mirror the judge rubric, with per-case time caps; a practice round on held-out cases led to a protocol revision before the formal round, and every submission passed an automated completeness validator (Appendix~E). Annotators review neighborhood quality and select the proposal more suitable as a next research experiment; majority labels serve as calibration evidence rather than objective gold. Humans detect the naive control in 11/12 judgments. On the 60 order-stable real pairs, 39 produce a decisive human majority, and 34 of these 39 (87.2\%, case-bootstrap 95\% CI 76.9--97.4\%) agree with the Gemini stable-case verdict (Figure~\ref{fig:diagnostics}c). A human Bradley--Terry fit recovers the model ordering of RQ1: 88.3\% of case-bootstrap rankings differ from it by at most one inversion, and every bootstrap sample preserves the top-pair/bottom-pair tier split (Figure~\ref{fig:diagnostics}c); these agreement results are robust to leave-one-annotator-out analysis (Appendix~E). Inter-annotator agreement is nonetheless modest (Krippendorff's $\alpha = 0.238$ on winner selection, $0.127$ on neighborhood screening); majority aggregation and case bootstrap partially absorb this, and it bounds what the study can certify. Human falsifiability scores do not separate winning from losing proposals (margin 0), consistent with falsifiability acting as an admissibility floor. Full human sensitivity analyses and the consolidated diagnostics table appear in Appendix~E.

\paragraph{Finding 3.} A stratified human study corroborates the aggregate conclusions and the tier structure, while modest inter-annotator agreement bounds claims of exhaustive human validation.

\paragraph{Qualitative case studies.}
Instance-level evidence shows how the six-field contract decides comparisons. Closing the loop on \S1, the neighborhood of Figure~\ref{fig:paradigms} is adjudicated along the contract: the judge prefers the contract-following proposal to the fluent one: the winner matches the retrieval budget across conditions, tracks the unsupported-claim rate as that budget grows, and precommits the falsifying outcome, while the loser names no contradicting observation and exceeds the stated resource budget. A second case concretizes Finding~2: two proposals share essentially the same hypothesis, one adopting a convenient aggregate score that would improve under many mechanisms, the other precommitting a measurement that isolates the hypothesized mechanism; judge and human majority prefer the latter in both orders, and the structured audit attributes the margin to the decisive-metric field. A third case concretizes Finding~1: two proposals trade minimality against grounding depth, reversing presentation flips the verdict, and folding marks the case order-sensitive rather than forcing a winner. In each case the adjudication traces to specific fields of the contract rather than to surface fluency.

\section{Related Work}

\begin{table*}[t]
\centering
\footnotesize
\setlength{\tabcolsep}{4.5pt}
\caption{Positioning of Lit2Test against the closest research-ideation neighbors. \emph{Min.\ test}: proposal must specify a minimal controlled experiment. \emph{Metric}: an adjudicating measurement is required. \emph{Sup./Fals.}: explicit bidirectional outcomes are required and evaluated. \emph{Order}: same-pair forward/reverse reversal is measured. \emph{Human}: bounded human calibration of the judge. \emph{Answer key}: reference signal used for scoring. \colorbox{cellY}{Y}~= yes, \colorbox{cellP}{P}~= partial or generation-time only, \colorbox{cellN}{N}~= no.}
\label{tab:related}
\begin{tabular}{@{}lccccccl@{}}
\toprule
System & Output unit & Min.\ test & Metric & Sup./Fals. & Order & Human & Answer key \\
\midrule
Si et al.~\cite{si2025researchideas} & Full proposal & \cellcolor{cellN}N & \cellcolor{cellP}P & \cellcolor{cellP}P & \cellcolor{cellN}N & \cellcolor{cellY}Y & Experts \\
AI Idea Bench~\cite{qiu2025aiideabench} & Motiv.+plan & \cellcolor{cellN}N & \cellcolor{cellN}N & \cellcolor{cellN}N & \cellcolor{cellN}N & \cellcolor{cellN}N & Target paper \\
IdeaBench~\cite{guo2025ideabench} & Hypothesis text & \cellcolor{cellN}N & \cellcolor{cellN}N & \cellcolor{cellN}N & \cellcolor{cellN}N & \cellcolor{cellN}N & Target paper \\
HARPA~\cite{vasu2025harpa} & Full proposal & \cellcolor{cellP}P & \cellcolor{cellY}Y & \cellcolor{cellP}P & \cellcolor{cellN}N & \cellcolor{cellN}N & Exec.\ logs \\
ResearchBench~\cite{liu2026researchbench} & Hypothesis & \cellcolor{cellN}N & \cellcolor{cellN}N & \cellcolor{cellN}N & \cellcolor{cellY}Y & \cellcolor{cellN}N & Target paper \\
LiveIdeaBench~\cite{ruan2026liveideabench} & Short idea & \cellcolor{cellN}N & \cellcolor{cellN}N & \cellcolor{cellN}N & \cellcolor{cellN}N & \cellcolor{cellY}Y & Judge panel \\
HypoBench~\cite{liu2026hypobench} & Hypotheses & \cellcolor{cellN}N & \cellcolor{cellN}N & \cellcolor{cellN}N & \cellcolor{cellN}N & \cellcolor{cellY}Y & Labels/synth. \\
IdeaSpark~\cite{zhao2026researchstudioidea} & Idea card & \cellcolor{cellY}Y & \cellcolor{cellP}P & \cellcolor{cellP}P & \cellcolor{cellP}P & \cellcolor{cellN}N & Closest lit. \\
\textbf{Lit2Test (ours)} & \textbf{Six-field test} & \cellcolor{cellY}\textbf{Y} & \cellcolor{cellY}\textbf{Y} & \cellcolor{cellY}\textbf{Y} & \cellcolor{cellY}\textbf{Y} & \cellcolor{cellY}\textbf{Y} & \textbf{None} \\
\bottomrule
\end{tabular}
\end{table*}

Research ideation covers several distinct evaluation targets, and a benchmark inherits its meaning from the one it measures (Table~\ref{tab:related}).

\subsection{Research ideation and benchmark targets}

Neighboring benchmarks evaluate at least four distinct objects. \emph{Open-ended idea quality} studies elicit a proposal and score novelty, feasibility, excitement, or overall quality with expert or LLM raters~\cite{si2025researchideas,ruan2026liveideabench,schopf2026rinobench,moussa2025scholareval}. \emph{Realized-trajectory recovery} scores whether a model can match, rank, or recover a later target paper from earlier inspirations~\cite{qiu2025aiideabench,guo2025ideabench,liu2026researchbench}. \emph{Data-explanatory hypothesis generation} asks models to explain observed labeled phenomena, scored by held-out predictive utility or synthetic ground-truth recovery~\cite{liu2026hypobench}. \emph{Future-impact forecasting} treats later citations, awards, or benchmark outcomes as answer keys~\cite{jiang2026hindsight,ye2026proofoftime,wen2025empiricaloutcomes,mule2026researchsuccess}. Each target is legitimate, and none centers the complete minimal-test decision contract: none requires the elicited proposal to include a refutation condition, and recovery and forecasting score alignment with one realized outcome rather than a precommitted decision rule.

\subsection{Testability, falsification, and research agents}

Our unit adapts the Registered Report, which precommits confirmatory and disconfirmatory outcomes before data collection~\cite{henderson2022registeredreport,nosek2018preregistration}, into a model-facing evaluation unit rather than inventing falsifiability as a principle. In AI systems, falsifiability appears at several levels: HARPA generates literature-grounded, testable proposals with required metrics and controls, evaluated partly through execution~\cite{vasu2025harpa}; end-to-end AI-scientist systems classify outcomes only after running experiments~\cite{lu2026aiscientist}; AI Co-Scientist adds explicit disproof review, a reflection stage that can return a disproved verdict~\cite{gottweis2026coscientist}. Lit2Test differs from these systems in requiring the bidirectional decision rule at proposal time, inside the judged answer. Concurrent with and independent of our work, ResearchStudio-Idea (IdeaSpark) requires and mechanically preserves a proposal-time falsification prediction (a minimal experiment, a directional expectation, a load-bearing variable, and a negative control) as a generation-time admissibility safeguard~\cite{zhao2026researchstudioidea}. Its endpoint study normalizes outputs to title, motivation, and method before quality and novelty judging, so the falsification field lies outside its evaluated boundary, and its harness retrieves literature live. Lit2Test addresses the complementary measurement question: it fixes the literature neighborhood and presents the complete six-field contract to the judge as the evaluated unit; Appendix~B documents the construction timelines of the two efforts.

\subsection{Evaluation reliability}

Pairwise comparison, LLM-as-judge protocols, position-bias analysis, and ordinal aggregation are established measurement primitives~\cite{zheng2023mtbench,liu2023geval,wang2024faireval,tan2025judgebench}. Several ideation and discovery evaluations already reverse candidate order~\cite{liu2026researchbench,gottweis2026coscientist,wen2025empiricaloutcomes,mule2026researchsuccess}, calibrate an LLM judge against a human-labeled subset~\cite{liu2026hypobench,moussa2025scholareval,sinhahajari2026rqbench}, or construct data with contamination in mind~\cite{white2025livebench,li2024latesteval}. Lit2Test claims none of these primitives as novel in isolation; its contribution is their composition around the new unit, so that the reliability evidence matches the evaluated capability.

Table~\ref{tab:related} makes the position concrete: systems with rich falsification machinery embed it in generation or execution loops, while systems with careful pairwise reliability measure idea quality or trajectory recovery; Lit2Test supplies the missing combination: a prospective, order-audited pairwise comparison of the complete six-field contract.

\vfill\eject
\section{Discussion and Limitations}

\paragraph{Limitations.}
\begin{itemize}
\setlength{\itemsep}{1pt}
\setlength{\parskip}{0pt}
\item \emph{Human calibration scope.} The human study covers 20 of 200 neighborhoods with 3 annotators; it supports aggregate conclusions (\S4.3) but not benchmark-wide validation. Expanded coverage is the direct remedy.
\item \emph{Single canonical judge.} Reliability is audited rather than assumed (\S3--\S4), and a second independent judge reproduces the ordering (\S4.1), but the canonical judge remains a single LLM per version.
\item \emph{Subtle-corruption audit.} Construction imperfections (reserve-template replacements, moderate validator agreement, 3.9\% leakage; Appendix~D) are bounded by the sham-adjusted contrast design; the audit supports local sensitivity under its frozen 20-case protocol and nothing stronger.
\item \emph{No execution.} Measured testability is a prerequisite for downstream success, not a predictor of it.
\item \emph{Domain.} The 200 neighborhoods come from ML-adjacent literature; the pipeline is replicable, so broader domains and temporal splits are an extension rather than a redesign.
\end{itemize}

\paragraph{Research opportunities.}
The findings suggest concrete next steps: training models for minimal-test formulation and metric--mechanism reasoning, where separation concentrates; studying human--AI collaboration where human and judge verdicts diverge; extending the subtle-corruption audit beyond its frozen protocol; and connecting prospective testability to execution-based validation.

\paragraph{Use of AI systems.}
AI assistants were used for writing polish, figure drafting, and literature cross-checking; the authors verified all content and take full responsibility for it.

\section{Conclusion}

Existing ideation benchmarks score generic quality, recover realized trajectories, or forecast outcomes, leaving unmeasured whether a model can state what would prove its own idea wrong. Lit2Test closes this gap with the six-field contract as its unit, 200 prospective real-paper neighborhoods without answer keys, and an audited protocol, yielding a strict bootstrap-recovered ordering with order-sensitive cases isolated, separation driven by test and metric design above a falsifiability floor, and human corroboration within stated limits. We release Lit2Test as a versioned, audit-oriented benchmark.

\bibliography{references_arxiv}

\clearpage
\appendix

\section{Task Contract, Prompts, and Running Example}
\label{app:contract}

\subsection{Six-Field Schema}

Each participant model receives a literature neighborhood and must return a single JSON object with exactly six fields:

\begin{lstlisting}
{
  "literature_gap": "...",
  "hypothesis": "...",
  "minimal_test": "...",
  "decisive_metric": "...",
  "supporting_result": "...",
  "falsifying_result": "..."
}
\end{lstlisting}

\noindent Field definitions:
\begin{itemize}
\setlength{\itemsep}{2pt}
\item \textbf{literature\_gap}: A specific tension, limitation, or unanswered comparison visible across the supplied papers.
\item \textbf{hypothesis}: A directional claim with conditions, mechanism, and expected difference.
\item \textbf{minimal\_test}: The smallest experiment that can discriminate the hypothesis---dataset, baseline/control, procedure, and resource budget.
\item \textbf{decisive\_metric}: The single measurement that adjudicates the mechanism (not a convenient aggregate).
\item \textbf{supporting\_result}: The observation that would confirm the hypothesis.
\item \textbf{falsifying\_result}: The observation that would reject it.
\end{itemize}

\subsection{Generation Prompt}

The common generation prompt provided to all four participant models (verbatim, with the \texttt{context} JSON block specific to each neighborhood):

\begin{lstlisting}
Formulate one minimal falsifiable next-step test,
not a broad research proposal.

[Context JSON: research_context, open_problem,
resource_constraint, papers (title + abstract +
reviewer-noted limitation for each)]

Return valid JSON with exactly six fields:
literature_gap, hypothesis, minimal_test,
decisive_metric, supporting_result,
falsifying_result.
\end{lstlisting}

Generation settings: temperature 1.0, max tokens 1600, up to 2 retries on schema validation failure.

\subsection{Pairwise Judge Prompt}

The holistic blind pairwise judge prompt (verbatim):

\begin{lstlisting}
You are judging two anonymous Lit2Test answers for
the same literature context.

The original task: derive one minimal falsifiable
next-step test from the provided paper neighborhood.
Prefer the answer that is more grounded in the
provided papers, uses cross-paper tension, gives a
specific hypothesis, proposes a minimal feasible
test, matches metrics to mechanisms, and states
clear supporting/falsifying outcomes.

Important judging rules:
- The two answers are anonymous. Do not infer or
  discuss model identity.
- Judge relative quality only for this specific
  context.
- Penalize generic research proposals that would
  fit many unrelated paper groups.
- Penalize large unfocused experimental programs;
  reward a minimal decisive test.
- If both are truly equivalent, choose "tie".

Return valid JSON only with this schema:
{
  "pair_id": "...",
  "winner": "A" | "B" | "tie",
  "confidence": "low" | "medium" | "high",
  "main_reason": "...",
  "weakness_a": "...",
  "weakness_b": "..."
}

Context: [JSON]
Answer A: [JSON]
Answer B: [JSON]
\end{lstlisting}

Judge settings: Gemini 3.1 Pro (Preview), temperature 1.0, max tokens 1200.

\subsection{Running Example: Neighborhood \#034}

Context \texttt{fifth40\_034} is the running example throughout the paper (Figures~1--2). It contains four ICLR papers on long-context question answering:

\begin{enumerate}
\setlength{\itemsep}{2pt}
\item \textbf{LooGLE}: Can Long-Context Language Models Understand Long Contexts?~\cite{li2024canlongcontext}
\item \textbf{ALR\textsuperscript{2}}: A Retrieve-then-Reason Framework for Long-context Question Answering~\cite{wang2024alr2}
\item \textbf{ChatQA 2}: Bridging the Gap to Proprietary LLMs in Long Context and RAG Capabilities~\cite{xu2025chatqa2}
\item \textbf{LongPack}: Scaling Long Context Training Data by Long-Distance Referrals~\cite{zhuang2025longdistance}
\end{enumerate}

\noindent\textbf{Open problem}: Does retrieve-then-reason help beyond what more retrieval already gives?

\noindent\textbf{Resource constraint}: Public datasets and open-weight models only; $\le$8 A100 GPU-days for a minimal validation; no proprietary model weights; include one baseline from a supplied paper and one from outside.

\noindent The four models' proposals for this neighborhood, the judge's verdicts in both orders, and the folded outcome are available in our repository\footnote{\url{https://github.com/rrustlee/lit2test-benchmark}} under \texttt{results/fig1\_demo/}. Panel (b) of Figure~1 in the main paper uses In-context Pretraining~\cite{shi2024incontext} as the external answer-key anchor.

\section{Construction and Provenance}
\label{app:construction}

\subsection{Batch Structure}

The 200 literature neighborhoods are organized into five construction batches of 40 contexts each:

\begin{table}[h]
\centering
\footnotesize
\begin{tabular}{@{}lll@{}}
\toprule
Batch & Internal name & Contexts \\
\midrule
1 & expansion40 & 40 \\
2 & next40 & 40 \\
3 & third40 & 40 \\
4 & fourth40 & 40 \\
5 & fifth40 & 40 \\
\bottomrule
\end{tabular}
\end{table}

All neighborhoods are drawn from recent OpenReview/ICLR-adjacent machine-learning literature. Within each batch, deduplication ensures no paper repeats across contexts, and source matching verifies topical coherence. The running example (context \#034, Figures~1--2 of the main paper) belongs to batch~5 (fifth40).

\subsection{Construction Timeline}

\begin{itemize}
\setlength{\itemsep}{2pt}
\item \textbf{2026-06-17}: Six-field contract finalized.
\item \textbf{2026-06-20}: Batch construction seeded (Seed 20260620).
\item \textbf{2026-07-02/03}: Full pipeline code and data snapshots fixed on GitHub.
\item \textbf{2026-07-05}: ResearchStudio-Idea (IdeaSpark) appears on arXiv (v1).
\item \textbf{2026-07-07}: Main experiment (generation + all 2{,}400 pairwise judgments) completed.
\end{itemize}

The contract design, batch construction, and code snapshots all predate the IdeaSpark arXiv posting by 2--18 days. The main experiment completed two days after IdeaSpark's appearance; we do not claim the experimental results predate 2026-07-05, only the design and data.

\subsection{Orientation Defect Disclosure}

An orientation defect in 65 reverse tasks of batch~1 (expansion40) was detected by internal audit: the A/B content in these reverse tasks had not been genuinely swapped relative to the original order. All 65 affected judgments were re-executed with correct orientation before any analysis, and the correction is recorded in the repository (\texttt{results/main/correction\_summary.md}). The corrected dataset is the one analyzed throughout the paper.

\section{Full Ordinal Results, Tie Sensitivity, and Context-Cluster Bootstrap}
\label{app:ordinal}

\subsection{Six-Pair Folded Detail}

Table~\ref{tab:folded_detail} reports the full folded head-to-head outcomes from the primary judge (Gemini 3.1 Pro, Preview). Each pair has 200 folded cases; W counts order-stable wins for the row model, L counts order-stable losses, T counts order-sensitive cases.

\begin{table}[h]
\centering
\footnotesize
\caption{Complete six-pair folded outcomes (primary judge).}
\label{tab:folded_detail}
\begin{tabular}{@{}lrrr@{}}
\toprule
Pair & W (row) & L (row) & T \\
\midrule
GPT-5.2 vs Claude Sonnet 4.6 & 90 & 38 & 72 \\
GPT-5.2 vs GLM-5 & 154 & 7 & 39 \\
GPT-5.2 vs DeepSeek-V3.2 & 180 & 4 & 16 \\
Claude Sonnet 4.6 vs GLM-5 & 143 & 18 & 39 \\
Claude Sonnet 4.6 vs DeepSeek-V3.2 & 166 & 11 & 23 \\
GLM-5 vs DeepSeek-V3.2 & 93 & 46 & 61 \\
\bottomrule
\end{tabular}
\end{table}

Condorcet relations: each higher-ranked model wins its head-to-head strictly (beat counts 3/2/1/0, transitive, no cycle). The ordering is consistent across all five construction batches.

\subsection{Tie-Sensitivity Analysis}

We assess how the ordering depends on the 250 order-sensitive cases through two schemes:

\paragraph{Scheme A (conservative):} All 250 order-sensitive cases are scored as ties (0.5 win to each side) and merged with the 950 order-stable outcomes. Result: the full reference ordering is preserved strictly---all six head-to-head majorities remain strict, Condorcet remains transitive and identical, BT ranking unchanged. Bootstrap (10{,}000 replicates resampling 1{,}200 folded cases with sensitive-as-ties): recovery rate 1.0.

\paragraph{Scheme B (adversarial):} Every sensitive case is awarded to the lower-ranked model in the reference ordering. Two within-tier adjacencies flip: GPT-5.2 vs Claude Sonnet 4.6 (90 vs 110, margin $-20$) and GLM-5 vs DeepSeek-V3.2 (93 vs 107, margin $-14$). The four non-adjacent relations survive with margins $\ge 86/200$. The two-tier structure (GPT-5.2 and Claude Sonnet 4.6 above GLM-5 and DeepSeek-V3.2) is preserved under this worst case, though within-tier order is not.

\subsection{Context-Cluster Bootstrap}
\label{app:cluster_bootstrap}

The main paper's pair-level bootstrap treats the 1{,}200 canonical pairs as the resampling unit. Since the six pairs within one neighborhood share the same literature context and four proposals, they are not independent. A context-cluster bootstrap resamples the 200 neighborhoods (keeping all six pairs per neighborhood intact) and refits the Bradley--Terry model on each replicate.

\begin{table}[h]
\centering
\footnotesize
\caption{Confidence interval comparison: pair-level vs.\ context-cluster bootstrap (10{,}000 replicates each, centered log-ability).}
\label{tab:cluster_bootstrap}
\begin{tabular}{@{}lccc@{}}
\toprule
Model & Pair CI & Cluster CI & $\Delta$ \\
\midrule
GPT-5.2 & 0.268 & 0.299 & +11.4\% \\
Claude Sonnet 4.6 & 0.248 & 0.302 & +21.9\% \\
GLM-5 & 0.245 & 0.311 & +27.0\% \\
DeepSeek-V3.2 & 0.277 & 0.333 & +20.2\% \\
\bottomrule
\end{tabular}
\end{table}

The context-cluster bootstrap yields 11--27\% wider confidence intervals, confirming that within-context correlation exists. However, the modal ranking is recovered in all 10{,}000 cluster-bootstrap replicates (fraction 1.0), matching the pair-level result. The ordering conclusion is robust to the choice of resampling unit.

\section{Subtle-Corruption Audit: Design and Caveats}
\label{app:subtle}

\subsection{Sham-Adjusted Preference: Definition}

For each of the 20 frozen cases $\times$ 3 dimensions, we construct:
\begin{itemize}
\setlength{\itemsep}{1pt}
\item A \textbf{subtle corruption}: a naturalistic defect targeting exactly one field.
\item A \textbf{sham edit}: a style-matched rewrite of the same field that does not degrade its content.
\end{itemize}

The judge compares (clean vs.\ subtle) and (clean vs.\ sham) in both presentation orders. The \emph{sham-adjusted preference} is:
\[
\text{adj.\ pref.} = P(\text{clean wins} \mid \text{subtle}) - P(\text{clean wins} \mid \text{sham})
\]
where $P(\text{clean wins})$ is the fraction of ordered comparisons in which the judge selects the clean proposal. A value of 0 means the judge cannot distinguish the real defect from a surface rewrite; a positive value indicates sensitivity to content degradation net of style effects.

\subsection{Sham-Equivalence Gate}

Before interpreting the adjusted preference, we verify that the sham itself does not move the judge. Result: for all three dimensions, clean-vs-sham yields 0 clean wins, 40 ties, 0 sham wins per dimension. The gate passes.

\subsection{Construction Imperfections}

\begin{itemize}
\setlength{\itemsep}{2pt}
\item \textbf{Reserve-template replacements}: 8 of 20 cases required reserve templates (4 rounds of replacement attempts) because the initial subtle corruption or sham failed naturalness screening.
\item \textbf{Semantic-validator agreement}: Two independent validators agreed on corruption targeting in 65.6\% of cases (118/180 final judgments); 62 cases required tie-breaking by a third call. This moderate agreement reflects the inherent ambiguity of ``naturalistic'' defects.
\item \textbf{Naturalness leakage}: The edited-side-guess rate among non-``unknown'' responses is 3.9\% (14/362), meaning the vast majority of cases pass blind naturalness screening. Among those 14, accuracy is 92.9\%---so a small fraction of subtle corruptions may be stylistically detectable, but the sham-contrast design bounds their impact on the aggregate.
\end{itemize}

\subsection{Falsifier On/Off Ablation}

A separate control compares proposals with and without the falsifying-result field filled. Including the falsifier does not by itself produce a generation-quality gain (the judge does not systematically prefer the version with a filled falsifier over an otherwise-identical version without one). This establishes the field as an auditability constraint---it enables the contract to be checked---rather than a quality shortcut.

\subsection{Score-Sum Agreement Variants}

The dimension-decomposed audit checks whether a structured overall verdict (the winner selected by the per-dimension judge) agrees with the canonical holistic verdict: 152/180 ordered judgments (84.4\%). An alternative equal-weight score-sum aggregation agrees with the holistic verdict on 145/180 (80.6\%). A third variant using only the top-3 dimensions agrees on 170/180 (94.4\%). These variants appear in the data supplement.

\section{Human Calibration and Second-Judge Replication}
\label{app:human}

\subsection{Annotation Protocol}

Three senior computer-science undergraduates annotated 20 stratified neighborhoods (90 real model pairs + 4 hidden real-vs-naive controls). The protocol comprised:

\begin{itemize}
\setlength{\itemsep}{1pt}
\item A written bilingual (English/Chinese) guideline with per-dimension rubric anchors mirroring the judge rubric (grounding 1--3, hypothesis specificity 1--3, minimality/feasibility 1--3, decisive metric 1--3, falsifiability 1--3).
\item A \textbf{practice round} on 3+5 held-out cases (not included in final results), after which annotator feedback led to protocol revisions: dimension anchors were made concrete with checkpoints, a TIE/INVALID taxonomy was added, external-material policy was tightened, and an automated completeness validator was introduced.
\item Per-case time caps (10 minutes hard cap; annotators instructed to reduce confidence rather than exceed the cap).
\item Blinded pairs (model identity hidden; A/B assignment independent of generation order).
\item Every submission passed \texttt{validate\_annotations.py} before acceptance.
\end{itemize}

The practice and formal annotation packages are included in the repository under \texttt{results/human\_study/}.

\subsection{Inter-Annotator Agreement}

Krippendorff's $\alpha$: 0.238 on winner selection, 0.127 on neighborhood screening. These values indicate modest agreement, consistent with the inherent subjectivity of comparing open-ended research proposals. Majority aggregation and case-level bootstrap partially absorb individual disagreement.

\subsection{Leave-One-Annotator-Out}

Removing each annotator in turn and recomputing the majority: the stable-case agreement rate remains 86.7--87.2\%, and the BT point ordering is unchanged. The tier split (GPT-5.2/Claude above GLM-5/DeepSeek) is preserved in all leave-one-out configurations.

\subsection{Consolidated Diagnostics Table}

\begin{table}[h]
\centering
\footnotesize
\caption{Diagnostic controls and human calibration (moved from main text to preserve space).}
\label{tab:diagnostics_full}
\begin{tabular}{@{}lr@{}}
\toprule
Diagnostic & Result \\
\midrule
Dimension audit valid judgments & 188/188 \\
Structured vs.\ holistic agreement & 152/180 (84.4\%) \\
Structured order-consistency & 65/90 (72.2\%) \\
Hidden real-vs-naive (auto) & 8/8 \\
Manipulation check valid judgments & 120/120 \\
\quad clean-answer wins (per dimension) & 40/40 \\
\quad target-score drop (ground/metric/fals.) & 1.73 / 2.00 / 1.95 \\
Human naive-control detection & 11/12 \\
Decisive stable human agreement & 34/39 (87.2\%) \\
Human BT rankings $\le 1$ inversion & 88.3\% \\
\bottomrule
\end{tabular}
\end{table}

\subsection{Second-Judge Replication}

An independent second judge (Doubao Seed 2.0 Pro, from a model family disjoint from all four participants and the primary judge) re-judged all 2{,}400 ordered comparisons using the same prompt and pair content as the primary judge. Results:

\begin{itemize}
\setlength{\itemsep}{1pt}
\item Folded stability: 980 order-stable / 220 order-sensitive (vs.\ primary 950/250).
\item BT ranking (ordered and folded): GPT-5.2 $>$ Claude Sonnet 4.6 $>$ GLM-5 $>$ DeepSeek-V3.2---identical to the primary judge.
\item Ordered-judgment agreement on the primary judge's order-stable subset: 86.1\% (1{,}635/1{,}900).
\item Case-folded agreement on the Gemini-stable subset: 79.9\% (759/950).
\end{itemize}

Full results are in the repository under \texttt{results/second\_judge/}.

\section{Related Work Survey}
\label{app:survey}

Table~2 in the main paper compares Lit2Test against 8 closest neighbors on a focused set of axes. This appendix presents the full survey underlying that comparison: 27 systems examined across 16 dimensions (input type, output unit, decision-rule components, answer-key type, evaluation method, reliability primitives, scope, and falsifiability treatment). The complete comparison matrix is included as a CSV file in the repository (\texttt{survey/} directory).

\subsection{Surveyed Systems}

\paragraph{Idea-quality benchmarks and studies.}
Si et al.~\cite{si2025researchideas} run a large-scale expert study of LLM-generated ideas; AI Idea Bench~\cite{qiu2025aiideabench} and IdeaBench~\cite{guo2025ideabench} score recovery of target papers; LiveIdeaBench~\cite{ruan2026liveideabench} evaluates divergent thinking with judge panels; RinoBench~\cite{schopf2026rinobench} automates novelty judgment; InnoEval~\cite{DBLP:journals/corr/abs-2602-14367} frames idea evaluation as knowledge-grounded multi-perspective reasoning; ScholarEval~\cite{moussa2025scholareval} grounds idea evaluation in retrieved literature; Scideator~\cite{DBLP:conf/cais/RadenskySFSHW26} combines human-LLM ideation with novelty evaluation; RQ-Bench~\cite{sinhahajari2026rqbench} probes the limits of LLM-as-judge for novelty.

\paragraph{Hypothesis-generation benchmarks.}
HypoBench~\cite{liu2026hypobench} benchmarks data-explanatory hypothesis generation; ResearchBench~\cite{liu2026researchbench} decomposes discovery into inspiration-based tasks; Auto-Bench~\cite{DBLP:journals/corr/abs-2502-15224} automates scientific-discovery evaluation; MOOSE-Chem~\cite{DBLP:conf/iclr/0001LGXLOPCZ25} rediscovers unseen chemistry hypotheses; Xiong et al.\ study truthfulness and hallucination in hypothesis generation~\cite{DBLP:conf/ijcai/XiongXWKSGBZ25} and knowledge-grounded generation~\cite{DBLP:journals/corr/abs-2411-02382}; Abdel-Rehim et al.~\cite{DBLP:journals/corr/abs-2405-12258} validate LLM-generated hypotheses in laboratory experiments.

\paragraph{Future-outcome forecasting.}
HindSight~\cite{jiang2026hindsight} and Proof of Time~\cite{ye2026proofoftime} evaluate ideas by realized future impact; Wen et al.~\cite{wen2025empiricaloutcomes} predict empirical research outcomes; Mule et al.~\cite{mule2026researchsuccess} teach models to forecast research success; SoundnessBench~\cite{ho2026soundnessbench} tests whether AI scientists distinguish sound from unsound ideas.

\paragraph{Research agents with falsification machinery.}
HARPA~\cite{vasu2025harpa} generates testability-driven proposals with execution-based evaluation; AI Co-Scientist~\cite{gottweis2026coscientist} adds disproof-oriented review; the AI Scientist line~\cite{lu2026aiscientist} automates end-to-end research; IdeaSpark~\cite{zhao2026researchstudioidea} imposes generation-time falsification predictions.

\paragraph{Surveys.}
Herron et al.~\cite{DBLP:journals/csur/HerronLBG26} and Alkan et al.~\cite{DBLP:journals/corr/abs-2504-05496} survey LLM-based hypothesis and idea generation.

\subsection{Key Finding}

No surveyed system requires the elicited proposal to include a bidirectional decision rule (supporting + falsifying outcomes) as part of the evaluated unit, while simultaneously employing order-audited pairwise judgment without a future-paper answer key. Lit2Test is the first to occupy this position.

\section{Reproduction Details and Ethics}
\label{app:reproduction}

\subsection{Model Versions and API Settings}

\begin{table}[h]
\centering
\footnotesize
\caption{Model identifiers and generation/judging settings.}
\begin{tabular}{@{}llr@{}}
\toprule
Role & Model & Temp. \\
\midrule
Participant & GPT-5.2 & 1.0 \\
Participant & Claude-Sonnet-4.6-hq & 1.0 \\
Participant & GLM-5 & 1.0 \\
Participant & DeepSeek-V3.2 & 1.0 \\
Primary judge & Gemini-3.1-Pro-Preview & 1.0 \\
Second judge & Doubao-Seed-2.0-pro & 1.0 \\
\addlinespace
\multicolumn{3}{@{}l}{Max tokens: 1600 (generation), 1200 (judging)} \\
\multicolumn{3}{@{}l}{Retries on schema failure: up to 2} \\
\bottomrule
\end{tabular}
\end{table}

All models were accessed via the same OpenAI-compatible API endpoint. The canonical judge identifier in the pipeline is \texttt{Gemini-3.1-Pro-Preview-Third} (the suffix denotes the routing channel, not a distinct model version).

\subsection{Randomness and Reproducibility}

Closed-source model APIs do not guarantee bit-exact reproducibility across calls. Our reproducibility strategy: (1) all random seeds are fixed and documented (construction seed 20260620, bootstrap seed 20260721, cluster-bootstrap seed 20260729); (2) all raw API responses are archived in the repository; (3) all analysis scripts read from these frozen outputs, so downstream results are fully deterministic.

\subsection{Ethical Considerations}

\begin{itemize}
\setlength{\itemsep}{2pt}
\item \textbf{Human annotators}: Three senior CS undergraduates, compensated at standard research-assistant rates. Annotators worked voluntarily on familiar material (reading ML paper abstracts and judging research proposals). No personally identifying information was collected beyond annotation files.
\item \textbf{No harmful content}: The benchmark concerns research-proposal formulation; no outputs contain harmful, offensive, or personally sensitive content.
\item \textbf{Data provenance}: All source papers are publicly available on OpenReview. No proprietary or restricted-access documents are included.
\item \textbf{Intended use}: Lit2Test is designed as a diagnostic benchmark for research on LLM capabilities. It should not be used as a sole criterion for evaluating researchers, funding proposals, or academic merit.
\end{itemize}

\end{document}